\pdfoutput=1

\documentclass[11pt]{article}

\usepackage{EMNLP2023}

\usepackage{times}
\usepackage{latexsym}

\usepackage{booktabs}
\usepackage{enumitem}

\usepackage[T1]{fontenc}

\usepackage[utf8]{inputenc}
\usepackage{microtype}
\usepackage{fancyhdr}
\usepackage{microtype}
\usepackage{multirow}
\usepackage{graphicx}
\usepackage{inconsolata}

\usepackage{lipsum}
\newcommand\blfootnote[1]{%
  \begingroup
  \renewcommand\thefootnote{}\footnote{#1}%
  \addtocounter{footnote}{-1}%
  \endgroup
}

\title{Offensive Language Identification \\ in Transliterated and Code-Mixed Bangla}

\author{Md Nishat Raihan\textsuperscript{1}, Umma Hani Tanmoy\textsuperscript{1}, Anika Binte Islam\textsuperscript{1}, Kai North\textsuperscript{1} \\\textbf{Tharindu Ranasinghe\textsuperscript{2}, Antonios Anastasopoulos\textsuperscript{1}, Marcos Zampieri\textsuperscript{1}}\\
        \textsuperscript{1}George Mason University, USA \\
        \textsuperscript{2}Aston University, UK \\
        \texttt{mraihan2@gmu.edu}}

\begin{document}
\maketitle
\begin{abstract}
Identifying offensive content in social media is vital for creating safe online communities. Several recent studies have addressed this problem by creating datasets for various languages. In this paper, we explore offensive language identification in texts with transliterations and code-mixing, linguistic phenomena common in multilingual societies, and a known challenge for NLP systems. We introduce TB-OLID, a transliterated Bangla offensive language dataset containing 5,000 manually annotated comments. We train and fine-tune machine learning models on TB-OLID, and we evaluate their results on this dataset. Our results show that English pre-trained transformer-based models, such as fBERT and HateBERT achieve the best performance on this dataset. 
\end{abstract}

\section{Introduction}

As the popularity of social media continues to grow, the spread of offensive content in these platforms has increased substantially, motivating companies to invest heavily in content moderation strategies and robust models to detect offensive content. We have observed a growing interest in this topic, evidenced by popular shared tasks at SemEval \cite{basile2019semeval} and other venues. Apart from a few notable exceptions \cite{mandl2020overview}, most of the work on this topic has not addressed the question of transliteration and code-mixing, two common phenomena in social media.\blfootnote{\bf WARNING: This paper contains examples that are offensive in nature.}

Code-mixing is the phenomenon of embedding linguistic units such as phrases, words, or morphemes of one language into another language \cite{myers1997duelling, muysken2000bilingual}. Code-mixed texts often feature transliterations where speakers use an alternative script to the language's official or standard script by mapping from one writing system (e.g., Hindi and its original Devanagari script) to another one (e.g., Latin transliteration of Hindi) based on phonetic similarity. Transliterated texts are widely used in social media platforms as transliteration allows users to write in their native language using a script that may not be supported by the platform and/or using Latin-based default keyboards. Furthermore, the use of transliteration also allows users to easily switch between languages with otherwise different scripts (e.g., English and Hindi). As discussed in a recent survey \cite{winata2022decades}, however, processing code-mixing datasets is a challenge that hinders performance in a variety of NLP tasks, thus deserving special attention.

Code-mixing and transliteration are common in various languages, including Bangla \cite{das2015code, jamatia2015part}. 
Related work on Bangla offensive language identification \cite{mridha2021boost}, however, has mostly focused on standard Bangla script. As such, the performance of offensive language identification models on code-mixing and transliterated Bangla remains largely unexplored. To address this shortcoming, we create TB-OLID, a manually annotated transliterated Bangla offensive language dataset. TB-OLID was annotated following the popular OLID hierarchical taxonomy \cite{OLID}, allowing cross-lingual experiments. To the best of our knowledge, the dataset is the first of its kind for Bangla, opening exciting new avenues for future research.

\begin{table*}[!ht]
\centering
\scalebox{0.9}{\begin{tabular}{p{14cm}ccc}
    \toprule
        \textbf{Comment} & \textbf{C|T} & \textbf{O|N} & \textbf{I|G|U} \\
        \midrule
        BN: \textit{Tui to 1 ta rastar chele mother chod.} &   \multirow{2}{*}{\textbf{C}} & \multirow{2}{*}{\textbf{O}} &\multirow{2}{*}{\textbf{I}} \\
        EN: You are a motherfucking street vagabond &  &  &  \\ \midrule
        BN: \textit{O to Manush na sokun} &   \multirow{2}{*}{\textbf{T}} & \multirow{2}{*}{\textbf{O}} &\multirow{2}{*}{\textbf{I}} \\
        EN: He/She is not a person but a vulture &  &  &  \\  \midrule
        BN: \textit{R kichudin por kanglu ra Uth ar mut diye cha banabe} &   \multirow{2}{*}{\textbf{T}} & \multirow{2}{*}{\textbf{O}} &\multirow{2}{*}{\textbf{G}} \\
        EN: After some days, the barbarians will make tea with camel piss &  &  &  \\	\midrule
        BN: \textit{Dhoren r dog ar baccha gulo ke gono dholai diye pongu kore den} &   \multirow{2}{*}{\textbf{C}} & \multirow{2}{*}{\textbf{O}} &\multirow{2}{*}{\textbf{G}} \\
        EN: Capture these son of bitches and beat them to their death &  &  &  \\    \midrule
        BN: \textit{Pagole kina bole chagole kina khai} &   \multirow{2}{*}{\textbf{T}} & \multirow{2}{*}{\textbf{O}} &\multirow{2}{*}{\textbf{U}} \\
        EN: A mad man and an animal have no difference &  &  &  \\ \midrule
        BN: \textit{Ami kintu parlam na hojom korte} &   \multirow{2}{*}{\textbf{T}} & \multirow{2}{*}{\textbf{N}} &\multirow{2}{*}{\textbf{}} \\
        EN: I cannot fathom it anymore &  &  &  \\        
    \bottomrule

\end{tabular}}
 \caption{Examples from TB-OLID in Bangla along with an English translation. The labels included are C (transliterated code-mixed), T (transliterated Bangla), O (offensive), N (not-offensive), I (offensive posts targeted at an individual), G (offensive posts targeted at a group), and U (untargeted offensive posts).} \label{tab:example_data} 
 \vspace{-1em}
\end{table*}

The main contributions of this paper are as follows:
\begin{enumerate}
    \item We introduce TB-OLID, an offensive language identification corpus containing 5,000 Facebook comments.\footnote{\url{https://github.com/LanguageTechnologyLab/TB-OLID}} 
    \vspace{-4mm}
   \item  We provide a comparative analysis of various machine learning models trained or fine-tuned on TB-OLID. 
\end{enumerate}

\section{Data}

\paragraph{Data Collection} We collect data from Facebook, the most popular social media platform in Bangladesh. We compile a list of the most popular Facebook pages in Bangladesh using Fanpage Karma\footnote{\url{https://www.fanpagekarma.com/}} and scraped comments from each of the top 100 most followed Facebook pages using the publicly available Facebook scraper tool.\footnote{\url{https://github.com/kevinzg/facebook-scraper}} This results in an initial corpus of over 100,000 comments. We exclude all comments not written with non-Latin script. We search the corpus using keywords for transliterated hate speech and offensive language. We select keywords from the list of 175 offensive Bangla terms by \citet{karim2021deephateexplainer}. As the dataset by \newcite{karim2020BengaliNLP} contains standard Bangla, we convert keywords into transliterated Bangla using the Indic-transliteration tool.\footnote{\url{https://github.com/sanskrit-coders/indic_transliteration_py}} Using these keywords we randomly select a set of 5,000 comments for annotation. 

\paragraph{Annotation Guidelines} We prepare the TB-OLID annotation guidelines containing labels and examples. The first step is to label whether a comment is transliterated Bangla or transliterated code-mixed Bangla. If the comment contains at least one English word along with other Bangla transliterated words, we consider it as transliterated code-mixed. Next, we consider the offensive vs. non-offensive distinction and, in the case of offensive posts, its target or lack thereof. Table \ref{tab:example_data} presents six annotated instances included in TB-OLID. 

We adopt the guidelines introduced by the popular OLID annotation taxonomy \cite{OLID} used in the OffensEval shared task \cite{offenseval} and replicated in multiple other datasets in languages such as Danish \cite{sigurbergsson2020offensive}, Greek \cite{pitenis2020offensive}, Marathi \cite{gaikwad2021cross, Zampieri2022}, Portuguese \cite{sigurbergsson2020offensive}, Sinhala \cite{ranasinghe2022sold} and Turkish \cite{coltekin2020}. We choose OLID due to the flexibility provided by its three-level hierarchical taxonomy that allows us to model different types of offensive and abusive content (e.g., hate speech, cyberbulling, etc.) using a single taxonomy. OLID's taxonomy considers whether an instance is offensive (level A), whether an offensive post is targeted or untargeted (level B), and what is the target of an offensive post (level C). As the second level of the TB-OLID annotation we consider OLID level A as follows.

\begin{itemize}[nolistsep,noitemsep]
    \item \textbf{Offensive:} Comments that contain any form of non-acceptable language or a targeted offense,  including insults, threats, and posts containing profane language
    \item \textbf{Non-offensive:} Comments that do not contain any offensive language 
\end{itemize}

\noindent Finally, the third level of the TB-OLID annotation merges OLIDs level B and C. We label whether a post is untargeted or, when targeted, whether it is labeled at an individual or a group as follows:

\begin{itemize}[noitemsep,nolistsep]
    \item \textbf{Individual:} Comments targeting any individual, such as mentioning a person with his/her name, unnamed participants, or famous personality. 
    \item \textbf{Group:} Comments targeting any group of people of common characteristics, religion, gender, etc. 
    \item \textbf{Untargeted:} Comments containing unacceptably strong language or profanities that are not targeted.
    
\end{itemize}


\paragraph{Ensuring Annotation Quality} Three annotators working on this project are tasked to annotate TB-OLID. They are PhD students in Computing aged 22-28 years old, 1 male and 2 female, all native speakers of Bangla and fluent speakers of English. The first step of the annotation process involves a pilot annotation study, where 300 comments are assigned to all three annotators to calculate initial agreement and refine the annotation guidelines according to their feedback. After this pilot experiment, we annotate an additional 4,700 Facebook comments totaling 5,000 instances which are subsequently split into 4,000 and 1,000 instances for training and testing, respectively. The instances in TB-OLID are annotated by at least two annotators, with the third one serving as adjudicator. We calculate pairwise inter-annotator agreement on 1,000 instances using Cohen's Kappa, and we report Cohen's Kappa score of 0.77 and 0.72 for levels 1 (code-mixed vs. transliterated) and 2 (offensive vs. non-offensive), which is generally considered substantial agreement. We report Cohen's Kappa score of 0.66 on level 3, considered moderate agreement. 


\paragraph{Dataset Statistics} We calculated the frequency of each label in the dataset namely code-mixed and transliterated, offensive and non-offensive, targeted and untargeted, and target types in the dataset. The dataset statistics are presented in Table \ref{tab:data_stat}.

\begin{table}[!ht]
\centering
\scalebox{0.95}{
\begin{tabular}{cccc}
\toprule
\bf Level & \bf Label & \bf Instances & \bf Percentage   \\ \midrule
\multirow{2}{*}{1} & T & 2,959 & 59.18\% \\ 
& C & 2,041 & 41.82\% \\ \midrule

\multirow{2}{*}{2} & O & 2,381 & 47.62\% \\
& N & 2,619 & 52.38\% \\  \midrule

\multirow{3}{*}{3} & I & 1,192 & 23.84\%  \\ 
 & G & 954 & 19.08\%  \\ 
& U & 235 & 4.70\%  \\ \bottomrule

\end{tabular}
}
\caption {TB-OLID per level and per class statistics. Percentage calculated considering the total number of instances in the dataset (5,000).} \label{tab:data_stat} 
\vspace{-1em}
\end{table}

\noindent Finally, we run an analysis of the code-mixed data using ad-hoc Python scripts. We observe that English is by far the most common language included in the code-mixed instances mixed with Bangla followed by Hindi. We report that 38.42\% of all tokens in the code-mixed (C) class are English. 

\section{Baselines and Models}

\paragraph{Baselines} We report the results of three baselines: (1) Google's Perspective API\footnote{\url{https://perspectiveapi.com/}}, a free API developed to detect offensive comments widely used as a baseline in this task \cite{10068619, fortuna2020toxic}; (2) prompting GPT 3.5 turbo providing the model with TB-OLID's annotation guidelines; and (3) a majority class baseline. Due to the API's limitations, Perspective API was used only for offensive language identification and not for for target classification. 

\paragraph{General Models} \label{llm} We experiment with pre-trained language models fine-tuned on TB-OLID. As our dataset is transliterated Bangla and contains English code-mixed, we experiment with BERT \cite{devlin2019bert}, roBERTa \cite{liu2019roberta} which are trained on English, and Bangla-BERT \cite{kowsher2022bangla}, which is trained on Bangla. We also use cross-lingual models such as mBERT \cite{devlin2019bert} and xlm-roBERTa \cite{conneau2020unsupervised} which are trained in multiple languages. 


\paragraph{Task-specific Models} We also experiment with task-specific fined-tuned models like HateBERT \cite{caselli2021hatebert}, and fBERT \cite{sarkar2021fbert}. These models were also further fined-tuned on TB-OLID.

\section{Results and Discussion}

We use F1-score to evaluate the performance of all models. The training and test sets are obtained by the aforementioned 80-20 random split on the entire TB-OLID dataset. We further subdivide the test set into transliterated code-mixed (C), transliterated (T), and all instances. We present results for offensive text classification (offensive vs. non-offensive) in Table \ref{tab:subtask_1}. 

\begin{table}[!ht]
\centering
\scalebox{0.95}{
\begin{tabular}{lccc}
\toprule
\textbf{Model} & \textbf{C} & \textbf{T} & \textbf{All}  \\
\midrule
fBERT  & 0.73  & 0.70 & 0.72 \\
HateBERT  & 0.74  & 0.69 & 0.72 \\
BERT  & 0.73  & 0.68  & 0.71 \\
m-BERT  & 0.70  & 0.68  & 0.69 \\
\textit{GPT 3.5} & 0.65  & 0.64  & 0.64 \\
\textit{Majority Class Baseline} &  0.57 & 0.57  & 0.57 \\
\textit{Perspective API}  & 0.53  & 0.50  & 0.51 \\
Bangla-BERT  & 0.42  & 0.42  & 0.42 \\
xlm-roBERTa  & 0.40  & 0.41  & 0.41 \\
roBERTa  & 0.41 & 0.41  & 0.41 \\
\bottomrule
\end{tabular}
}
\caption{Offensive Language Identification - F1-score of all models trained and/or fine-tuned on TB-OLID. We report results on the transliterated code-mixed (C), transliterated (T), and All test set. Baselines in italics.}
\label{tab:subtask_1}
\vspace{-1em}
\end{table}

We observe that the standard BERT model performs well over the baselines, whereas the Bangla-BERT model performs less well. BERT achieves F1-score of 0.71, whereas Bangla-BERT obtains F1-score of 0.42. We believe this is due to the fact that many instances in the dataset are in Latin script, which means that BanglaBERT frequently struggles with out-of-vocabulary tokens. The low performance of Bangla-BERT in this task requires further examination. Models pre-trained specifically on offensive language identification perform very well with fBERT and Hate-BERT coming out on top, both with an F1 score of 0.72. Finally, we observe that the top-5 performing models perform better on the code-mixing data compared to transliterated data. This is likely due to the heavy presence of English words in the code-mixing data where we observe the presence of 38\% of English words. 

Finally, Table \ref{tab:subtask_2} presents the results of target type classification (individual, group, or untargeted). 

\begin{table}[ht]
\centering
\scalebox{0.95}{
\begin{tabular}{lccc}
\toprule
\textbf{Model} & \textbf{C} & \textbf{T} & \textbf{All} \\
\midrule
HateBERT & 0.69  & 0.66 & 0.68 \\
m-BERT & 0.72  & 0.64  & 0.67 \\
BERT & 0.72 & 0.64 & 0.67 \\
fBERT  & 0.66  & 0.64  & 0.65 \\
roBERTa & 0.73  & 0.60  & 0.65 \\
\textit{GPT 3.5} & 0.39  & 0.46  & 0.43 \\
\textit{Majority Class Baseline} & 0.48  & 0.63 & 0.55 \\
xlm-roBERTa  & 0.61 & 0.51  & 0.55 \\
Bangla-BERT  & 0.59  & 0.47 & 0.51 \\
\bottomrule
\end{tabular}
}
\caption{Target Classification - F1-score of all models trained and/or fine-tuned on TB-OLID. We report results on the transliterated code-mixed (C), transliterated (T), and All test sets. Baselines in italics.}
\label{tab:subtask_2} 
\vspace{-1em}
\end{table}

\noindent Overall, target classification is a more challenging task than offensive language identification due to the presence of three classes instead of two. Therefore, all results are substantially lower for this task. HateBERT performs better than all other models with an F1 score of 0.68. roBERTa achieved more competitive performance for target classification than for offensive language identification whereas Bangla-BERT did not perform well in both tasks. Finally, similar to the previous task, the best-performing models achieved higher F1 scores on the code-mixed data than on the transliterated data. 

One key observation is that the transformer-based models do not perform very well, since most of them are not pre-trained on transliterated Bangla. Among the models that we experiment with, only xlm-roBERTa is pre-trained with a comparatively small set of Romanized Bangla. However, the lack of any standard rules for spelling in transliterated Bangla makes TB-OLID very challenging.

\section{Conclusion and Future Work}

In this work, we introduced TB-OLID, a transliterated Bangla offensive language dataset containing 5,000 instances retrieved from Facebook. Three native speakers of Bangla have annotated the dataset with respect to the presence of code-mixing, the presence of offensive language, and its target according to the OLID taxonomy. TB-OLID opens exciting new avenues for research on offensive language identification in Bangla.  

We performed experiments with multiple models such as general monolingual models like BERT \cite{devlin2019bert}, roBERTa \cite{liu2019roberta} and Bangla-BERT \cite{kowsher2022bangla}; cross-lingual models like mBERT \cite{devlin2019bert} and xlm-roBERTa \cite{conneau2020unsupervised}; and models fine-tuned for offensive language identification like HateBERT \cite{caselli2021hatebert}, and fBERT \cite{sarkar2021fbert}). The best results were obtained by the task-specific models.

In future work, we would like to extend the TB-OLID dataset and annotate the offense type (e.g., religious offense, political offense, etc.). This would help us identify the common targets in various platforms. Furthermore, we would like to pre-train and fine-tune a Bangla transliterated BERT model to see how it performs on TB-OLID. Finally, in future work, we would like to evaluate the performance of other recently released large language models (LLMs) (e.g., GPT 4.0, Llama 2) on TB-OLID. The first baseline results using GPT 3.5 indicate that general-purpose LLMs still struggle with the transliterated and code-mixed content presented in TB-OLID.


\section*{Acknowledgments}

We thank the anonymous workshop reviewers for their insightful feedback. Antonios Anastasopoulos is generously supported by NSF award IIS-2125466.

\section*{Ethics Statement}
The generation and annotation procedure of TB-OLID adheres to the \href{https://www.aclweb.org/portal/content/acl-code-ethics}{ACL Ethics Policy} and seeks to make a valuable contribution to the realm of online safety. The technology in question possesses the potential to serve as a beneficial instrument for the moderation of online content, thereby facilitating the creation of safer digital environments. However, it is imperative to exercise caution and implement stringent regulations to prevent its potential misuse for purposes such as monitoring or censorship.


\bibliography{clean}
\bibliographystyle{acl_natbib}

\end{document}